\documentclass[conference]{IEEEtran}
\usepackage[T1]{fontenc}

\usepackage{xcolor}
\usepackage{amsbsy}
\usepackage{amssymb}
\usepackage{graphicx}

\ifCLASSINFOpdf

\else

\fi

\usepackage{multirow}
\usepackage{float}
\usepackage[export]{adjustbox}
\usepackage{xcolor}
\usepackage{afterpage}
\usepackage[labelsep=period, figurename=Figure]{caption}
\usepackage[caption=false,font=footnotesize]{subfig} 
\usepackage{amsmath}
\graphicspath{{figure/}}
\usepackage{verbatim}
\usepackage{graphicx}
\usepackage[caption=false,font=footnotesize]{subfig} 
\usepackage{tikz}
\usepackage{textcomp}
\usepackage{lipsum}
\usepackage{comment}
\usepackage{algpseudocode}
\usepackage{amsmath}
\usepackage{algorithm}
\usepackage{cite}
\usepackage{graphicx}
\usepackage{textcomp}
\usepackage{xcolor}
\usepackage{amsmath,amssymb,amsfonts}
\usepackage{booktabs}
\usepackage{enumitem}
\setlist[enumerate,1]{label={\arabic*.}}
\makeatother
 \usepackage{authblk}
\IEEEoverridecommandlockouts
\usepackage{fancyhdr}
\usepackage{epstopdf}
\usepackage{authblk}
\usepackage[utf8]{inputenc}
\DeclareUnicodeCharacter{2212}{-}

\author{
Omar Erak,~\IEEEmembership{Member,~IEEE},
Omar Alhussein,~\IEEEmembership{Senior Member,~IEEE},
Fang Fang,~\IEEEmembership{Senior Member,~IEEE}, \\ and Sami Muhaidat,~\IEEEmembership{Senior Member,~IEEE}

\thanks{Omar Erak, Omar Alhussein and Sami Muhaidat are with the KU 6G Research Center,  College of Computing and Mathematical Sciences, Khalifa University, Abu Dhabi, UAE (e-mail: omarerak@ieee.org, omar.alhussein@ku.ac.ae).}

\thanks{Fang Fang is with the Department of Electrical and Computer Engineering, and also with the Department of Computer Science, Western University, London, ON N6A 3K7, Canada (e-mail: fang.fang@uwo.ca).}
}

\begin{document}
\raggedbottom
\addtolength{\textfloatsep}{-2.19pt}

%

\title{Topology-Preserving Deep Joint Source-Channel Coding for Semantic Communication}

\maketitle


\begin{abstract}
    Many wireless vision applications, such as autonomous driving, require preservation of global structural information rather than only per-pixel fidelity. However, existing Deep joint source-channel coding (DeepJSCC) schemes mainly optimize pixel-wise losses and provide no explicit protection of connectivity or topology. This letter proposes TopoJSCC, a topology-aware DeepJSCC framework that integrates persistent-homology regularizers to end-to-end training. Specifically, we enforce topological consistency by penalizing Wasserstein distances between cubical persistence diagrams of original and reconstructed images, and between Vietoris--Rips persistence of latent features before and after the channel to promote a robust latent manifold. TopoJSCC is based on end-to-end learning and requires no side information. Experiments show improved topology preservation and peak signal-to-noise ratio (PSNR) in low  signal-to-noise ratio (SNR) and bandwidth-ratio regimes.

\end{abstract}

\begin{IEEEkeywords} Semantic communication, deep joint-source channel coding, persistent
homology, topological data analysis
\end{IEEEkeywords}

\IEEEpeerreviewmaketitle

\section{Introduction}

Next-generation wireless systems are expected to support safety-critical and perception-centric applications such as autonomous driving and telemedicine \cite{wang2023road}. In these scenarios, the structure of the received visual data is often more important than exact pixel-wise fidelity: a broken road segment can mislead a path planner, and a missing branch in a retinal vascular tree may invalidate a diagnosis. Small topological errors can cause downstream semantic modules to fail even when traditional signal-level metrics such as peak signal-to-noise ratio (PSNR) remain high, highlighting the need for communication schemes that explicitly protect topology and other task-relevant semantics. 

Conventional wireless image transmission typically follows the separation principle, combining source compression with error-control coding and modulation. While spectrally efficient, such systems are sensitive to channel mismatch and suffer the ``cliff effect,'' where reconstruction quality collapses once the signal-to-noise ratio (SNR) drops below the design point. Deep joint source-channel coding (DeepJSCC) instead learns an end-to-end mapping from pixels to complex channel symbols using encoder/decoder networks with a differentiable channel layer, yielding graceful degradation and strong performance in low-SNR or bandwidth-limited regimes \cite{bourtsoulatze2019deep}. However, existing DeepJSCC methods optimize only pixel-wise or geometric distortions and may fail to preserve global structure, so reconstructions of structured images such as roads or retinal vascular trees can still contain broken, missing, or incorrectly connected segments.

Topological data analysis (TDA), in particular persistent homology (PH), provides compact descriptors of multi-scale connectivity and loop structure through persistence diagrams \cite{edelsbrunner2002topological, tdafrontiers}. TDA has shown promising results in computer vision \cite{hu2019topology, moor2020topological} and recently, TDA has been introduced into communication systems. TopoCode \cite{10921659} encodes persistence-diagram information as side information for error detection and correction in image communication, transmitting a compact topological signature alongside conventional data. While effective at correcting semantically significant errors at low SNR with limited redundancy, such schemes operate on top of classical digital coding and do not exploit TDA inside an end-to-end learned physical layer.



To the best of our knowledge, persistent-homology-based losses have not been integrated directly into DeepJSCC for wireless image transmission, nor has topology preservation been systematically evaluated in this context. This leaves a gap between topology-aware perception and learned physical-layer schemes, especially for applications where the topology of thin structures, such as road networks, is critical. This letter addresses this gap with the following contributions:
\begin{itemize}
\item We propose TopoJSCC, a topology-aware DeepJSCC framework that augments the mean squared error (MSE) distortion with an image-domain persistent-homology loss based on a Wasserstein distance (WD) between persistence diagrams of original and reconstructed images.
\item We introduce a latent-space topological loss that applies PH to Vietoris--Rips complexes built from encoder features before and after the channel, promoting a channel-robust latent manifold without changing the architecture.
\item We evaluate TopoJSCC on two topology-rich datasets under additive white Gaussian noise (AWGN) and Rayleigh fading, comparing against DeepJSCC, BPG+LDPC, and a TopoCode-based baseline, and show that TopoJSCC substantially reduces persistence-diagram distortion and topological errors while maintaining PSNR comparable to DeepJSCC.
\end{itemize}

\enlargethispage{\baselineskip} 

\begin{figure}
    \centering
    \includegraphics[width=0.85\linewidth]{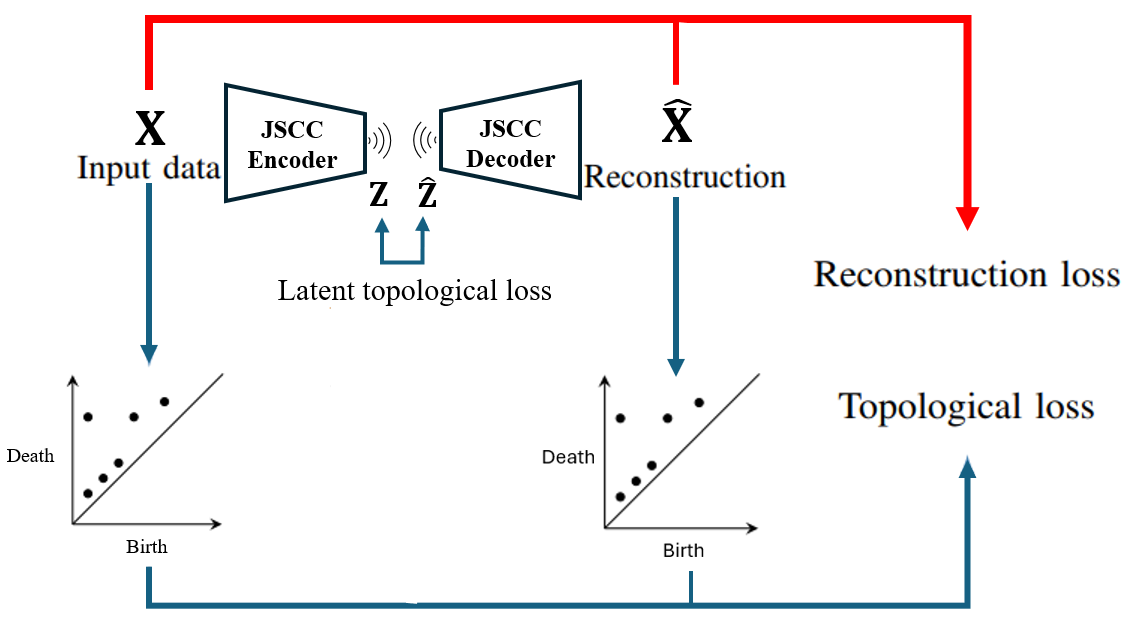}
    \caption{TopoJSCC system model and losses}
    \label{fig:placeholder}
\end{figure}

    
    \section{System Model}
    
    We consider a point-to-point wireless semantic communication system designed for topology-sensitive image data, such as road networks or retinal vasculature. The source image is represented by $\mathbf{X} \in \mathbb{R}^{H \times W \times C}$, where $H$ and $W$ denote height and width, and $C$ is the number of channels. For processing, $\mathbf{X}$ is reshaped into a vector $\mathbf{x} \in \mathbb{R}^{n}$ with $n = HWC$. 
    
    The transmitter employs a semantic encoder, modeled as a deterministic mapping $f_\theta: \mathbb{R}^{n} \to \mathbb{R}^{2k}$ parameterized by a deep neural network (DNN) with weights $\theta$. Given $\mathbf{x}$, the encoder produces $\mathbf{s} = f_\theta(\mathbf{x}) \in \mathbb{R}^{2k}$, which is grouped into $k$ complex channel symbols $z_\ell = s_{2\ell-1} + j s_{2\ell}$, $\ell = 1,\dots,k$, forming the complex channel input vector $\mathbf{z} \in \mathbb{C}^{k}$. Unlike conventional separation-based schemes, $f_\theta$ jointly performs source and channel coding. We impose an average power constraint normalized by the number of channel uses, $\frac{1}{k}\,\mathbb{E}[\|\mathbf{z}\|_2^2] \le P$, where the expectation is taken over the source distribution. The channel bandwidth ratio is defined as $\rho \triangleq k/n$, and we focus on the bandwidth-limited regime $\rho < 1$.
    
    The encoded symbol vector $\mathbf{z}$ is transmitted over a complex baseband wireless channel. The received signal $\mathbf{y} \in \mathbb{C}^{k}$ is given by $
        \mathbf{y} = h\,\mathbf{z} + \mathbf{n},$
    where $h \in \mathbb{C}$ is the complex channel gain and $\mathbf{n} \sim \mathcal{CN}(\mathbf{0}, N_0 \mathbf{I}_k)$ denotes AWGN. We consider two channel environments: a static AWGN channel with $h = 1$, and a slow Rayleigh fading channel with $h \sim \mathcal{CN}(0,1)$, which is constant for the duration of one image transmission and varies independently between images. A normalization layer enforces the power constraint. With $\mathbb{E}[|h|^2]=1$, the average receive signal-to-noise ratio is $\mathrm{SNR} = P/N_0$, and we report $\mathrm{SNR}_{\mathrm{dB}} = 10\log_{10}(P/N_0)$.
    
    At the receiver, a semantic decoder $g_\phi: \mathbb{R}^{2k} \to \mathbb{R}^{n}$, parameterized by a DNN with weights $\phi$, processes the noisy signal. In implementation, the real and imaginary parts of $\mathbf{y}$ are concatenated into a real vector $\tilde{\mathbf{y}} \in \mathbb{R}^{2k}$, and the decoder outputs $\hat{\mathbf{x}} = g_\phi(\tilde{\mathbf{y}})$, which is reshaped back to $\hat{\mathbf{X}} \in \mathbb{R}^{H \times W \times C}$ for evaluation. 
    

\section{Proposed Topology-Aware DeepJSCC}

Topology-rich images, such as road networks or vascular structures, are
characterized by their connectivity, branching structure, and loop patterns. Standard
pixel-wise losses such as MSE do not explicitly enforce preservation of these
structures. To address this, we augment the DeepJSCC objective with two
 PH-based regularizers: an image-domain topological loss
that matches the topology of the input and reconstructed images, and a
latent-space topological loss that constrains the topology of the latent
representations before and after the channel. PH characterizes how the homology of a space changes along a
filtration (i.e., nested sequence of cell complexes). Each filtration yields a
persistence diagram, a multiset of birth–death pairs
$(b_i^m, d_i^m)$, indicating when a $m$-dimensional feature appears and
disappears \cite{tdafrontiers}. We compare diagrams using a $p$-Wasserstein distance \cite{kaczynski2006computational}. For both losses we compute
PH in dimensions $m \in \{0,1\}$, where $m=0$ corresponds to
connected components, and $m=1$ to one-dimensional cycles.

\subsection{Image-Domain Topological Loss}

For the image-domain loss, we  represent the $H \times W$ pixel grid as a cubical complex, and denote by
$Q_{ij} = [i,i+1] \times [j,j+1]$ the unit square associated with pixel
$(i,j)$, together with its edges and vertices. The union of these cells forms a
cubical complex $\mathcal{K}(\mathbf{X})$. The image intensities define values
$X_{ij} \in [0,1]$ on the pixels, which we identify with the squares $Q_{ij}$. To emphasize bright foreground structures, we adopt a superlevel-set
filtration. For a threshold $\tau \in [0,1]$, we define
\begin{equation}
    \mathcal{X}_\tau
    = \bigcup_{i,j :\, X_{ij} \ge \tau} Q_{ij}.
    \label{eq:superlevel}
\end{equation}
As $\tau$ is decreased, more pixels enter
the complex. In implementation we consider a finite, strictly decreasing
sequence of thresholds $1 = \tau_0 > \tau_1 > \dots > \tau_T = 0,$ which yields the nested filtration $\emptyset
    \subseteq \mathcal{X}_{\tau_0}
    \subseteq \mathcal{X}_{\tau_1}
    \subseteq \dots
    \subseteq \mathcal{X}_{\tau_T}
    = \mathcal{K}(\mathbf{X}).
    \label{eq:image-filtration}
$
Persistent homology applied to this filtration tracks the creation and
destruction of connected components and loops in the bright regions of the
image.

We denote the resulting persistence diagrams by
\[
    \mathcal{D}^{\text{img}}_{m}(\mathbf{X})
    = \mathrm{PH}_m\big(\{\mathcal{X}_{\tau_t}\}_{t=0}^{T}\big),
    \quad m \in \{0,1\},
\]
and analogously obtain $\mathcal{D}^{\text{img}}_{m}(\hat{\mathbf{X}})$ from
the reconstructed image $\hat{\mathbf{X}}$ using the same threshold sequence. To quantify topological discrepancies, we compare persistence diagrams using a $p$-Wasserstein distance. Let $D$ and $D'$ be two persistence diagrams (multisets of points in
$\mathbb{R}^2$). We augment each diagram with the diagonal
$\Delta = \{(t,t) : t \in \mathbb{R}\}$ of infinite multiplicity so that
unmatched points can be paired with the diagonal. Let
$\eta : D \cup \Delta \rightarrow D' \cup \Delta$ be a bijection. For
$p \ge 1$, the $p$-Wasserstein distance between $D$ and $D'$ is
\begin{equation}
    d_{W,p}(D, D')
    =
    \left(
        \inf_{\eta}
        \sum_{x \in D \cup \Delta}
        \|x - \eta(x)\|_\infty^{p}
    \right)^{1/p},
    \label{eq:wasserstein}
\end{equation}
where for a point $x=(a,b)$ we use the $\ell_\infty$ norm
$\|(a,b)\|_\infty = \max\{|a|,|b|\}$. Intuitively, $\eta$ matches birth–death
pairs in $D$ to those in $D'$, and pairs that cannot be matched are mapped to
the diagonal. In all our experiments we fix $p=2$ and write $d_{W,2}$. Compared with the bottleneck distance, which is dominated by the single largest mismatch between diagrams, the 2-Wasserstein distance aggregates discrepancies over all features and provides a more informative signal when used as a regularizer in our DeepJSCC training.

\enlargethispage{\baselineskip} 
For each homology dimension $m \in \{0,1\}$, we measure the discrepancy between
the topologies of $\mathbf{X}$ and $\hat{\mathbf{X}}$ as
\[
    d^{\text{img}}_{m}\big(\mathbf{X}, \hat{\mathbf{X}}\big)
    = d_{W,2}\big(\mathcal{D}^{\text{img}}_{m}(\mathbf{X}), 
                  \mathcal{D}^{\text{img}}_{m}(\hat{\mathbf{X}})\big),
\]
and define the image-domain topological loss
\[
    L_{\text{top}}^{\text{img}}(\mathbf{X}, \hat{\mathbf{X}})
    = \sum_{m \in \{0,1\}}
      d^{\text{img}}_{m}\big(\mathbf{X}, \hat{\mathbf{X}}\big).
\]
Minimizing $L_{\text{top}}^{\text{img}}$ encourages the reconstruction to
preserve the connected components and loops of the bright structures in the
image, which is crucial for road networks and retinal vasculature.

\subsection{Latent-Space Topological Loss}

We also promote robustness of the latent representation to channel noise.
Consider a mini-batch of $B$ images $\{\mathbf{X}_i\}_{i=1}^{B}$ with vectorized forms $\mathbf{x}_i \in \mathbb{R}^{n}$.
The encoder $f_\theta$ produces real latent vectors before the channel, namely $\mathbf{s}_i = f_\theta(\mathbf{x}_i) \in \mathbb{R}^{2k}$ for $i = 1,\dots,B$.
Let $\mathcal{C}_{\mathrm{r}}(\cdot)$ denote the differentiable channel layer in its real-valued implementation; it maps $\mathbf{s}_i$ to the stacked real and imaginary parts of the received complex channel output.
The noisy latent vectors are then $\tilde{\mathbf{s}}_i = \mathcal{C}_{\mathrm{r}}(\mathbf{s}_i) \in \mathbb{R}^{2k}$. We interpret the sets $ = \{\mathbf{s}_1,\dots,\mathbf{s}_B\}, \quad
    \tilde{S} = \{\tilde{\mathbf{s}}_1,\dots,\tilde{\mathbf{s}}_B\} $
as point clouds in $\mathbb{R}^{2k}$ endowed with the Euclidean distance.

To summarize the geometry of these point clouds, we construct a Vietoris--Rips
filtration \cite{tdafrontiers}. Given a finite metric space $(S,d)$, the Vietoris–Rips complex at scale
$\varepsilon \ge 0$ is the abstract simplicial complex
\[
    \mathcal{R}_\varepsilon(S)
    = \big\{ \sigma \subseteq S \,\big|\,
        \sigma \text{ finite and }
        \max_{u,v \in \sigma} d(u,v) \le \varepsilon
      \big\}.
\]
We consider a strictly increasing sequence of scales $0=\varepsilon_0<\varepsilon_1<\dots<\varepsilon_T=\varepsilon_{\max}$, which yields the Vietoris–Rips filtration
    $\mathcal{R}_{\varepsilon_0}(S)
    \subseteq \mathcal{R}_{\varepsilon_1}(S)
    \subseteq \dots
    \subseteq \mathcal{R}_{\varepsilon_T}(S),$
using the Euclidean metric $d(u,v) = \|u-v\|_2$ on $\mathbb{R}^{2k}$. Applying
PH to this filtration produces, for $m \in \{0,1\}$,
persistence diagrams
    $\mathcal{D}^{\text{lat}}_{m}(S)
    = \mathrm{PH}_m\big(\{\mathcal{R}_{\varepsilon_t}(S)\}_{t=0}^{T}\big)
    \quad$ and
    $\mathcal{D}^{\text{lat}}_{m}(\tilde{S})
    = \mathrm{PH}_m\big(\{\mathcal{R}_{\varepsilon_t}(\tilde{S})\}_{t=0}^{T}\big).$ Using the same $2$-Wasserstein distance, we quantify the topological distortion
introduced by the channel via
\[
    d^{\text{lat}}_{m}(S, \tilde{S})
    = d_{W,2}\big(\mathcal{D}^{\text{lat}}_{m}(S),
                  \mathcal{D}^{\text{lat}}_{m}(\tilde{S})\big),
\]
and define the latent-space topological loss as follows
\[
    L_{\text{top}}^{\text{lat}}(S, \tilde{S})
    = \sum_{m \in \{0,1\}}
      d^{\text{lat}}_{m}(S, \tilde{S}).
\]
Minimizing $L_{\text{top}}^{\text{lat}}$ encourages the encoder to learn
latent manifolds whose global shape (clusters, branches, and loops) is
preserved after the stochastic channel perturbation.

\subsection{Final Training Loss}

For a mini-batch of $B$ image pairs
$\{(\mathbf{X}_i, \hat{\mathbf{X}}_i)\}_{i=1}^{B}$ with corresponding
vectorized forms $(\mathbf{x}_i, \hat{\mathbf{x}}_i)$ and latent sets
$(S,\tilde{S})$ constructed from the whole batch as above, we train the
topology-aware DeepJSCC by minimizing the batch objective

\begin{equation}
\begin{split}
    L_{\text{batch}} &= \frac{1}{B} \sum_{i=1}^{B}
    \left[ 
        L_{\text{MSE}}(\mathbf{x}_i, \hat{\mathbf{x}}_i)
        + \lambda_{\text{img}}
        L_{\text{top}}^{\text{img}}(\mathbf{X}_i, \hat{\mathbf{X}}_i)
    \right] \\
    &\quad + \lambda_{\text{lat}} L_{\text{top}}^{\text{lat}}(S, \tilde{S}),
\end{split}
\label{eq:batch-loss}
\end{equation}
where
\[
    L_{\text{MSE}}(\mathbf{x}_i, \hat{\mathbf{x}}_i)
    = \frac{1}{n}\,\|\mathbf{x}_i - \hat{\mathbf{x}}_i\|_2^2.
\]

\noindent The first term is an average over samples in the batch; the latent-space term
is naturally defined at the batch level because the persistence diagrams are
computed from the point clouds $S$ and $\tilde{S}$ that contain all $B$
latent codes. In practice, $B$ is fixed and the relative contribution of the
latent-space term is controlled by the hyperparameter $\lambda_{\text{lat}}$.
Gradients of both topological losses are obtained through differentiable
PH layers \cite{gabrielsson2020topology, hu2019topology} and are backpropagated jointly with the MSE term to update the encoder and decoder parameters. Using MSE as the base distortion keeps the training objective and the evaluation metric (PSNR, which is derived from MSE) aligned across all experiments similar to prior works \cite{bourtsoulatze2019deep}.

\subsection{Choice of Topological Hyperparameters}

The additional topological terms in the training objective operate on different numerical scales than the MSE reconstruction loss. To give them a meaningful but not dominating influence, we choose the weights $\lambda_{\text{img}}$ and $\lambda_{\text{lat}}$ in two stages.

We first train a baseline DeepJSCC model using only the MSE loss. After convergence, we freeze its parameters and evaluate, on a held-out validation batch, the typical magnitudes of the reconstruction loss and of the unweighted image- and latent-domain topological losses. We then set initial values for $\lambda_{\text{img}}$ and $\lambda_{\text{lat}}$ so that, on this batch, each weighted topological term contributes only a small fraction of the reconstruction loss. This provides a starting point where topology affects the gradients without overwhelming the reconstruction objective.

Starting from these initial weights, we perform a short grid search over nearby values on a validation set, selecting the pair $(\lambda_{\text{img}}, \lambda_{\text{lat}})$ that best trades off PSNR  and topological fidelity (smaller Wasserstein distances between input and reconstruction diagrams). To further stabilize training, we apply an annealing schedule
\[
    \lambda_{\text{img}}(t) 
    = \lambda_{\text{img}} \big(1 - e^{-t / T}\big), \quad
    \lambda_{\text{lat}}(t) 
    = \lambda_{\text{lat}} \big(1 - e^{-t / T}\big),
\]
where $t$ is the training epoch index and $T$ is a time constant. This gradually increases the influence of the topological losses, allowing the network to first learn a coarse reconstruction and then refine topology. In practice, we find that the image-domain coefficient can be chosen larger than the latent-space coefficient, since the batch-level latent loss may otherwise dominate the gradients. We find that setting $\lambda_{\text{img}} = 10^{-4}$ and $\lambda_{\text{lat}} =10^{-5}$, provides the most favourable results.

\begin{figure*}[t]
\centering
\subfloat[PSNR vs. SNR (AWGN)]{%
  \includegraphics[width=0.24\linewidth]{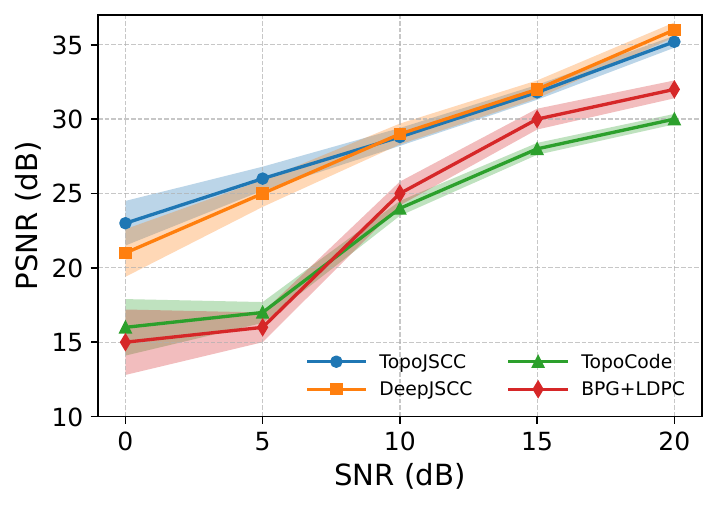}%
  \label{fig:omn_psnr_snr_awgn}}%
\hfil
\subfloat[PSNR vs. SNR (Rayleigh)]{%
  \includegraphics[width=0.24\linewidth]{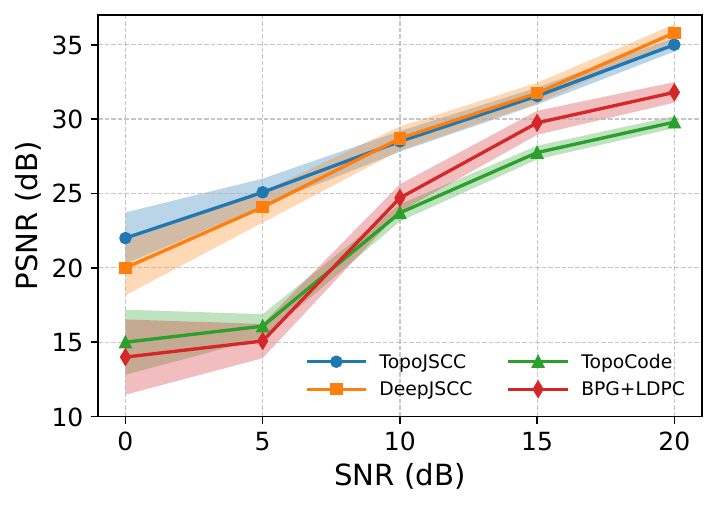}%
  \label{fig:omn_psnr_snr_ray}}
\subfloat[Wdist vs. SNR (AWGN)]{%
  \includegraphics[width=0.24\linewidth]{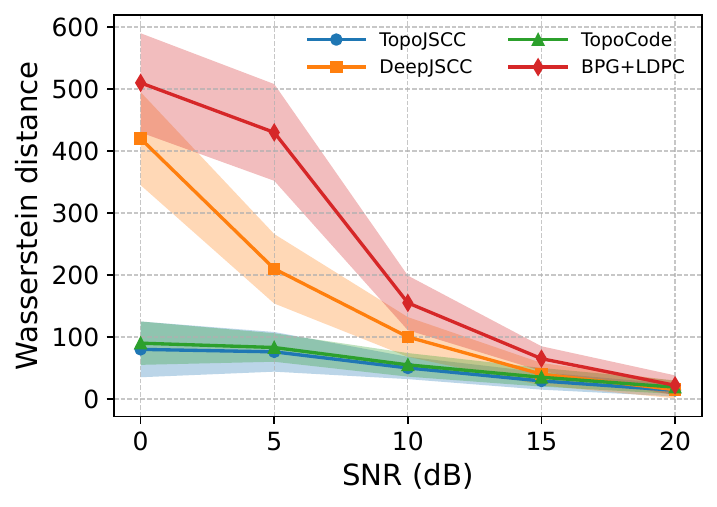}%
  \label{fig:omn_wdist_snr_awgn}}%
\hfil
\subfloat[Wdist vs. SNR (Rayleigh)]{%
  \includegraphics[width=0.24\linewidth]{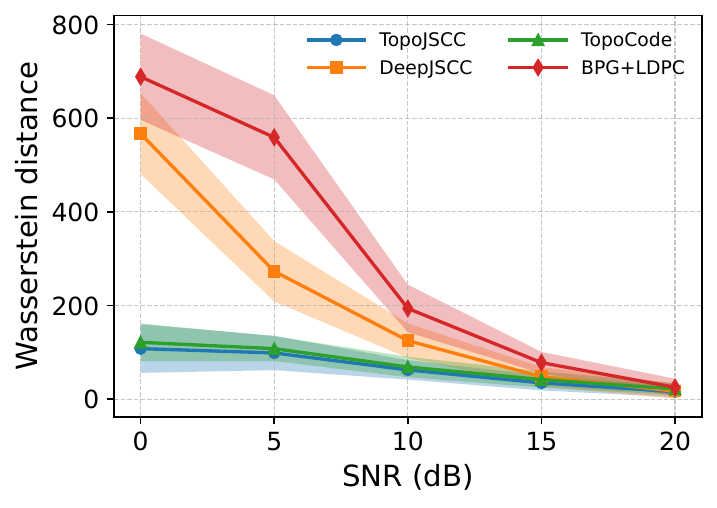}%
  \label{fig:omn_wdist_snr_ray}}%

\caption{Omniglot results under an SNR sweep. Bandwidth ratio is fixed at 0.40 and Topology-regularized DeepJSCC exhibits the most pronounced improvements in the low-SNR regime under both AWGN and Rayleigh fading. }
\label{fig:omniglot_snr}
\end{figure*}

\section{Results and Discussion}

\subsection{Experimental Setup}

\textbf{Datasets.} We evaluate on two topology-rich datasets: Omniglot \cite{lake2015human}, with 1623 handwritten characters from 50 alphabets, and the DeepGlobe Road Extraction dataset \cite{DeepGlobe18}, consisting of binary road-segmentation masks from satellite imagery. We represent all inputs as single-channel continuous images normalized to $[0,1]$, and train the JSCC models with an MSE reconstruction loss, aligning the training objective with the PSNR metric and enabling a unified treatment of Omniglot and road masks.

\textbf{JSCC Architecture and Channel.}
All neural JSCC schemes use the same CNN autoencoder backbone as in \cite{bourtsoulatze2019deep}. The encoder applies input normalization followed by five $5{\times}5$ convolutional layers with PReLU activations (strides $2,2,1,1,1$). The final convolution outputs a length-$2k$ real vector, which is power-normalized and grouped into $k$ complex channel symbols, so that the bandwidth ratio is $\rho = k/n$ with $n = HWC$. A non-trainable differentiable channel layer is inserted between encoder and decoder. The decoder mirrors the encoder with five transpose-convolution layers, using PReLU after each layer except the last; a final sigmoid produces the reconstructed image. For each bandwidth ratio $\rho$, we train a separate encoder-decoder pair; our topology-aware method differs only in the training loss.

\textbf{Comparison Schemes.}
We compare the proposed TopoJSCC with the original CNN-based DeepJSCC scheme \cite{bourtsoulatze2019deep} implemented with the same architecture but trained only with MSE, a classical separation-based scheme using the BPG image codec \cite{bellard2015bpg} followed by 5G LDPC codes \cite{richardson2018design} and TopoCode \cite{10921659}, a topology-aware digital error detection and correction scheme. 

\textbf{Training Protocol.}
All neural models are implemented in PyTorch and trained from scratch using the Adam optimizer with learning rate $10^{-4}$ and batch size of $128$. 
For each dataset and bandwidth ratio $\rho$, we employ an early stopping strategy to ensure convergence without overfitting. During training, the instantaneous channel SNR is randomly sampled for each mini-batch from a discrete set covering low-to-high SNRs ($\{0,5,10,15,20\}$ dB), and the same SNR distribution is used for all JSCC-based methods to enable SNR-adaptive behavior without retraining.  We note that the PH computations introduce additional training-time cost, but they do not affect the inference-time complexity, as the deployed encoder/decoder do not perform any topological calculations.


\textbf{Evaluation Metrics.}
At test time, we evaluate performance under two complementary sweeps: (1) SNR sweep at fixed bandwidth, where we fix the bandwidth ratio $\rho$ and vary the channel SNR, and (2) bandwidth sweep at fixed SNR, where we fix the SNR and vary $\rho$ across the considered compression ratios.  In the bandwidth sweep we consider $\rho \in \{0.05, 0.10, 0.20, 0.30, 0.40, 0.50\}$, corresponding to increasing numbers of channel uses.
For each operating point, we report the average PSNR between the input and reconstructed images, and the average $2$-Wasserstein distance between the persistence diagrams of the input and reconstructed images, computed on cubical complexes with a fixed threshold grid and homology dimensions $m \in \{0,1\}$. PSNR measures pixel-wise fidelity, while the Wasserstein distance quantifies preservation of the underlying topological structure. All results are averaged over multiple independent runs, with the shaded region indicating variability across 10 different runs.

\begin{figure*}[t]
\centering
\subfloat[PSNR vs. BW ratio (AWGN)]{%
  \includegraphics[width=0.24\linewidth]{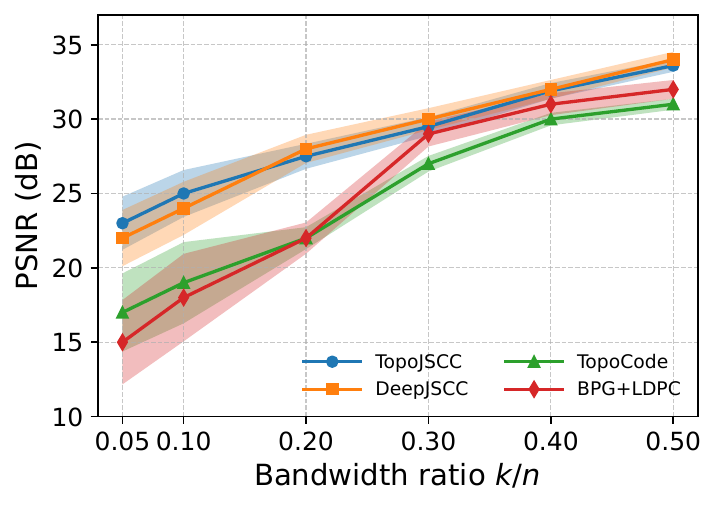}%
  \label{fig:roads_psnr_bw_awgn}}%
\hfill
\subfloat[PSNR vs. BW ratio (Rayleigh)]{%
  \includegraphics[width=0.24\linewidth]{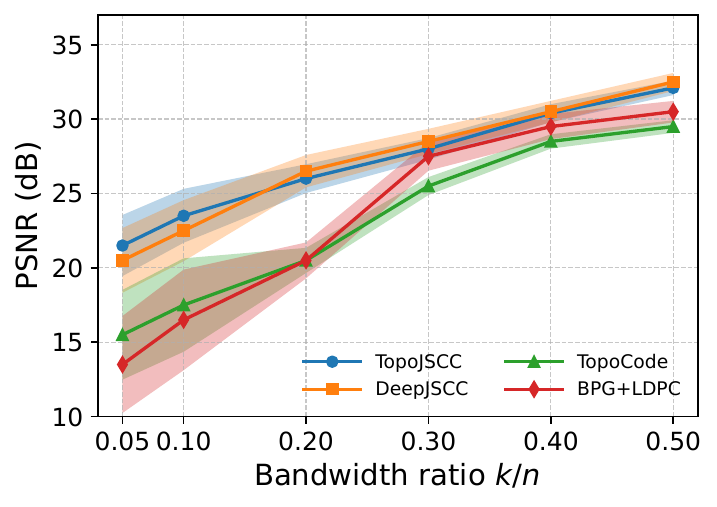}%
  \label{fig:roads_psnr_bw_ray}}%
\hfill
\subfloat[Wdist vs. BW ratio (AWGN)]{%
  \includegraphics[width=0.24\linewidth]{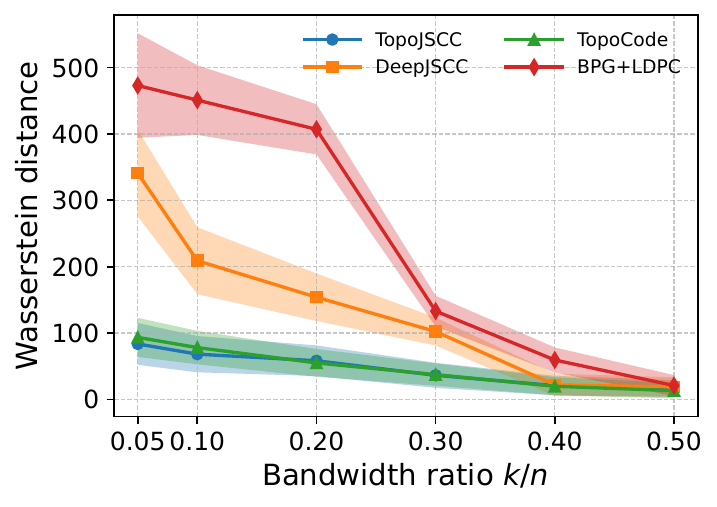}%
  \label{fig:roads_wdist_bw_awgn}}%
\hfill
\subfloat[Wdist vs. BW ratio (Rayleigh)]{%
  \includegraphics[width=0.24\linewidth]{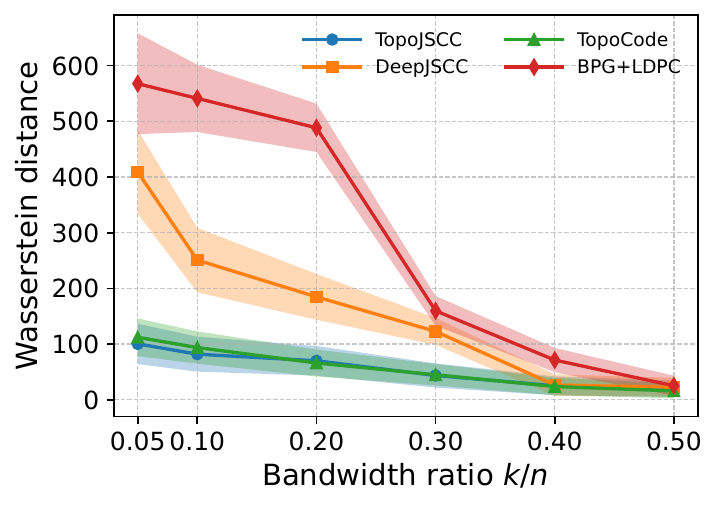}%
  \label{fig:roads_wdist_bw_ray}}%

\caption{DeepGlobe Road Extraction
dataset results under a bandwidth-ratio sweep. SNR is fixed at 15 dB. Topology-regularized DeepJSCC yields larger topology improvements at low bandwidth ratios under both AWGN and Rayleigh fading.}
\label{fig:roads_bw}
\end{figure*}

\begin{figure}[!h]
    \centering
    \includegraphics[width=0.89\columnwidth]{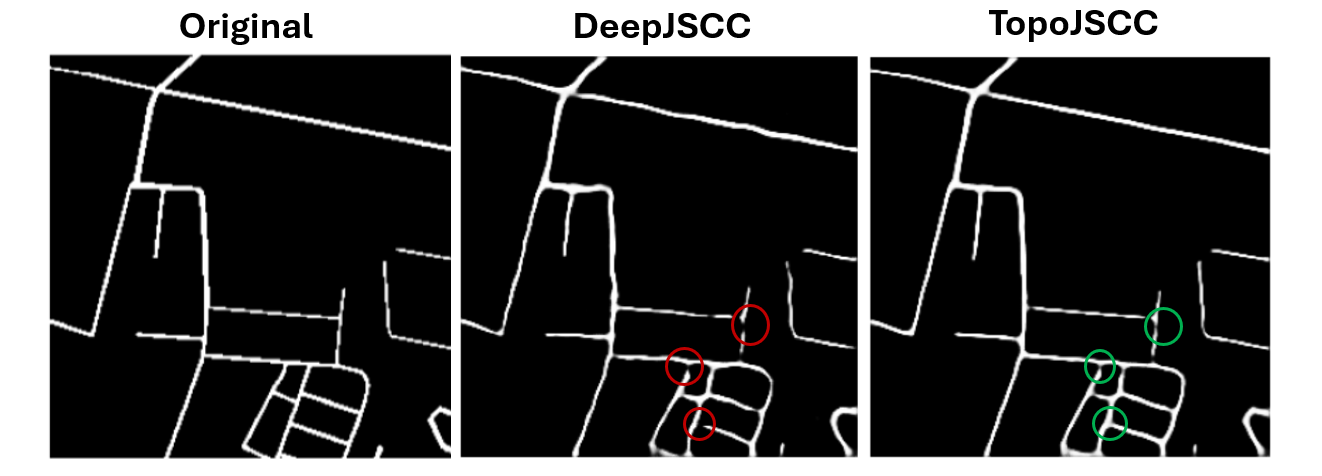}
    \caption{Reconstructions showing topological preservation at SNR = 5 dB.}
    \label{fig:recons}
\end{figure}

\subsection{Experimental Results}

Fig.~\ref{fig:omniglot_snr} reports the Omniglot results under an SNR sweep for both AWGN and Rayleigh channels. In terms of PSNR, the proposed TopoJSCC consistently outperforms the purely MSE-trained DeepJSCC at low SNR and remains within about $0.5$~dB of DeepJSCC at high SNR for both channel models. For example, at $\mathrm{SNR}=0$~dB over AWGN Fig.~\ref{fig:omniglot_snr}(a), TopoJSCC achieves roughly $2$~dB gain over DeepJSCC and $7$--$8$~dB over TopoCode and BPG+LDPC, mainly due to the cliff effect. As the SNR increases, all schemes improve and the gap between them narrows, but TopoJSCC consistently outperforms TopoCode and nearly matches the performance of DeepJSCC at a high SNR. The benefit of the proposed method becomes more pronounced in the topological metric. At low and moderate SNR, TopoJSCC yields a substantially smaller WD between persistence diagrams compared to DeepJSCC and BPG+LDPC, indicating better preservation of the underlying topology. At $\mathrm{SNR}=0$~dB over AWGN Fig.~\ref{fig:omniglot_snr}(c), the WD of TopoJSCC is roughly four to five times lower than that of DeepJSCC and BPG+LDPC, while remaining close to that of TopoCode. TopoCode attains very small WDs but at the cost of significantly degraded PSNR, whereas TopoJSCC offers a more favorable tradeoff between pixel fidelity and topological consistency. Similar trends are observed under Rayleigh fading Figs.~\ref{fig:omniglot_snr}(b),(d), confirming that the topology-aware loss improves robustness to channel variations without sacrificing reconstruction quality.

Fig.~\ref{fig:roads_bw} shows the results on the DeepGlobe road extraction data as a function of the bandwidth ratio $\rho=k/n$. Under both AWGN and Rayleigh channels, the PSNR curves Figs.~\ref{fig:roads_bw}(a),(b)  demonstrate that TopoJSCC remains competitive with DeepJSCC and BPG+LDPC across all ratios, despite the additional topological regularization. At very low bandwidth ratios TopoJSCC attains the highest PSNR among all schemes in both channel conditions, highlighting the benefit of topology-aware features in the heavily compressed regime. As $\rho$ increases, the performance of all schemes converges and the PSNR gap between TopoJSCC and DeepJSCC becomes negligible, while BPG+LDPC and TopoCode remain slightly below.

Figs.~\ref{fig:roads_bw}(c),(d), report the WD as a function of $\rho$. For all rates and both channel models, TopoJSCC significantly reduces the topological distortion compared with DeepJSCC and BPG+LDPC. At $\rho=0.05$, the WD for TopoJSCC is again several times lower than for the digital baseline and the MSE-only DeepJSCC, indicating markedly better preservation of road connectivity and loop structure in the reconstructed maps. TopoCode achieves one of the smallest distances as well, but its PSNR degradation relative to TopoJSCC underlines the advantage of joint semantic and topological optimization over purely topology-protected digital transmission. Qualitative examples further illustrate the structural benefits of TopoJSCC. In Fig.~\ref{fig:recons}, reconstructions produced by TopoJSCC exhibit fewer broken connections and holes compared to DeepJSCC under low SNR or bandwidth. These observations are consistent with the quantitative trends in Figs.~\ref{fig:omniglot_snr} and~\ref{fig:roads_bw}.  Finally, the ablation in Table~\ref{tab:ablation} shows that the image-domain topological loss provides the main reduction in persistence-diagram distortion, while the latent-space loss offers a smaller but complementary gain and combining both yields the best overall results.


\begin{table}[!h]
\centering
\caption{Ablation on Omniglot (SNR=0 dB, $\rho=0.4$).}
\label{tab:ablation}
\footnotesize
\begin{tabular*}{0.85\columnwidth}{@{\extracolsep{\fill}}lcc@{}}
\hline
Method & PSNR $\uparrow$ & Wdist $\downarrow$ \\
\hline
DeepJSCC & 22.1 & 415 \\
+$L_{\rm top}^{\rm img}$ only & 22.9 & 132 \\
+$L_{\rm top}^{\rm lat}$ only & 22.4 & 271 \\
Full TopoJSCC & \textbf{23.2} & \textbf{94} \\
\hline
\end{tabular*}
\end{table}

\enlargethispage{\baselineskip} 
\section{Conclusion}
This letter introduced TopoJSCC, a topology-aware DeepJSCC framework that integrates persistent-homology regularizers in both the image domain and the latent space to better preserve connectivity and loop structure in wireless image transmission. Experiments on topology-rich datasets under AWGN and Rayleigh fading show that TopoJSCC consistently reduces persistence-diagram distortion and achieves superior PSNR in low-SNR and low-bandwidth regimes, achieving a more favorable tradeoff than MSE-only DeepJSCC and digital baselines like TopoCode.

\bibliographystyle{IEEEtran}

\bibliography{references}





\end{document}